\documentclass[10pt,twocolumn,letterpaper]{article}

\usepackage{cvpr}              

\usepackage{graphicx}
\usepackage{amsmath}
\usepackage{amssymb}
\usepackage{booktabs}
\usepackage{multirow}
\usepackage{bigstrut}
\usepackage[table]{xcolor}
\usepackage[labelformat=simple]{subcaption}
\usepackage[accsupp]{axessibility}
\usepackage{balance}

\usepackage[pagebackref,breaklinks,colorlinks]{hyperref}

\usepackage[capitalize]{cleveref}
\crefname{section}{Sec.}{Secs.}
\Crefname{section}{Section}{Sections}
\Crefname{table}{Table}{Tables}
\crefname{table}{Tab.}{Tabs.}

\def\cvprPaperID{5642} 
\def\confName{CVPR}
\def\confYear{2023}

\begin{document}

\title{LG-BPN: Local and Global Blind-Patch Network for \\Self-Supervised Real-World Denoising}

\author{Zichun Wang $^{1}$, Ying Fu $^{1}$\thanks{Corresponding Author} , Ji Liu $^{2}$, Yulun Zhang $^{3}$\\
$^{1}$ Beijing Institute of Technology, $^2$ Baidu Inc., Beijing, China, $^3$ ETH Z\"{u}rich
\\{\tt\small \{wangzichun, fuying\}@bit.edu.cn, liuji04@baidu.com, yulun100@gmail.com}
}
\maketitle

\begin{abstract}

Despite the significant results on synthetic noise under simplified assumptions, most self-supervised denoising methods fail under real noise due to the strong spatial noise correlation, including the advanced self-supervised blind-spot networks (BSNs). For recent methods targeting real-world denoising, they either suffer from ignoring this spatial correlation, or are limited by the destruction of fine textures for under-considering the correlation. In this paper, we present a novel method called LG-BPN for self-supervised real-world denoising, which takes the spatial correlation statistic into our network design for local detail restoration, and also brings the long-range dependencies modeling ability to previously CNN-based BSN methods. First, based on the correlation statistic, we propose a densely-sampled patch-masked convolution module. By taking more neighbor pixels with low noise correlation into account, we enable a denser local receptive field, preserving more useful information for enhanced fine structure recovery. Second, we propose a dilated Transformer block to allow distant context exploitation in BSN. This global perception addresses the intrinsic deficiency of BSN, whose receptive field is constrained by the blind spot requirement, which can not be fully resolved by the previous CNN-based BSNs. These two designs enable LG-BPN to fully exploit both the detailed structure and the global interaction in a blind manner. Extensive results on real-world datasets demonstrate the superior performance of our method. 
\url{https://github.com/Wang-XIaoDingdd/LGBPN}
\end{abstract}

\vspace{-5mm}

\begin{figure}
    \centering
     \scriptsize
    \setlength{\tabcolsep}{0.05cm}
   
	\hspace{-3mm}
   \begin{minipage}[l]{0.302\linewidth}
    \vspace{-6.3mm}
\begin{center}
   \includegraphics[width=1.2\linewidth]{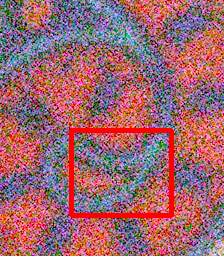}\vspace{0.35mm}
\\  \hspace{5mm}
\vspace{-6mm}
\begin{tabular}[c]{@{}c@{}}SIDD  Validation: \\ 0018-0031 \end{tabular}
\end{center}
    \end{minipage}
    \hspace{4.4mm}
    \begin{minipage}[t]{0.6\linewidth}
     \begin{tabular}{ccc}
      \includegraphics[width=0.33\linewidth]{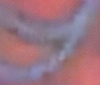}
     & \includegraphics[width=0.33\linewidth]{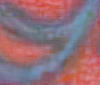}
     & \includegraphics[width=0.33\linewidth]{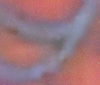}
     \\
     DnCNN \cite{zhang2017beyond}       & C2N \cite{jang2021c2n} &  R2R\cite{pang2021recorrupted} \\
      31.17/0.778 & 28.09/0.706 & 30.37/0.770\\
     \includegraphics[width=0.33\linewidth]{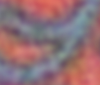}
       &\includegraphics[width=0.33\linewidth]{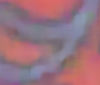}
         &\includegraphics[width=0.33\linewidth]{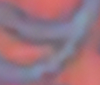}
      
        \\ CVF-SID \cite{neshatavar2022cvf}& AP-BSN \cite{lee2022ap}& \textbf{LG-BPN (Ours)}\\ 
        28.56/0.792 & 31.92/0.826 & \textbf{{32.76}}/\textbf{{0.897}}

   \\
     \end{tabular}
    \end{minipage}
   
    \vspace{-2mm}
    \caption{ Visual comparison of various methods on the SIDD validation \cite{abdelhamed2018high} dataset.
    Compared with DnCNN \cite{zhang2017beyond}, C2N \cite{jang2021c2n} and R2R \cite{jang2021c2n}, LG-BPN can be trained in a self-supervised manner without extra data.
    CVF-SID \cite{neshatavar2022cvf} still contains noise in the output, and AP-BSN \cite{lee2022ap} suffers from the loss of details.
    }
       \vspace{-5mm}
    \label{fig:sidd_val} 
   \end{figure}

\vspace{-1mm}
\section{Introduction}
\vspace{-1mm}

Image denoising is a fundamental research topic for low-level vision \cite{cheng2021nbnet,zhang2017beyond}. Noise can greatly degrade the quality of the captured images, thus bringing adverse impacts on the subsequent downstream tasks \cite{liu2020connecting,xie2015joint}.
Recently, with the rapid development of neural networks, 
learning-based methods have shown significant advances compared with traditional model-based algorithms \cite{buades2005non,dabov2007image,gu2014weighted,10003653}.

Unfortunately, learning-based methods often rely on massive labeled image pairs for training \cite{anwar2019real,yu2019deep,zamir2022restormer}.
This can not be simply addressed by synthesizing additive white Gaussian noise (AWGN) pairs, since the gap between AWGN and real noise distribution severely degrades their performance in the real world \cite{anwar2019real,guo2019toward}.
To this end, several attempts have been made for collecting real-world datasets \cite{abdelhamed2018high,brummer2019natural}. 
Nonetheless, its application is still hindered by the rigorously-controlled and labor-intensive collection procedure. For instance, capturing ground truth images requires long exposure or multiple shots, which is unavailable in complex situations, \textit{e.g.}, dynamic scenes with motion.

To alleviate the constraint of the large-scale paired dataset, methods without the need for ground truth have attracted increasing attention. 
The pioneer work Noise2Noise (N2N) \cite{lehtinen2018noise2noise} uses paired noisy observations for training, 
which can be applied when clean images are not available.
Still, obtaining such noisy pairs under the same scene is less feasible. 
To make self-supervised methods more practical, researchers seek to learn from one, instead of pairs of observations.
Among these methods, blind-spot networks (BSNs) \cite{krull2019noise2void,batson2019noise2self,wang2022blind2unblind,laine2019high} show significant advances
 to restore clean pixels by utilizing neighbor pixels,
with a special blind spot receptive field requirement.
Despite their promising results on simple noise such as AWGN, these methods usually work under simplified assumptions, \textit{e.g.}, the noise is pixel-wise independent. This obviously does not hold for real noise, where the distribution can be extremely complex and present a strong spatial correlation.

Accordingly, a few methods have been proposed for self-supervised real noise removal. 
Recorrupted-to-Recorrupted (R2R) \cite{pang2021recorrupted} tries to construct noisy-noisy pairs,
while it can not be directly applied without
extra information,
which is not practical in real situations.
CVF-SID \cite{neshatavar2022cvf} 
disentangles the noise components from noisy images,
but it assumes the real noise is spatially invariant and ignores the spatial correlation, which contradicts real noise distribution.

Recently, AP-BSN \cite{lee2022ap} combines pixel-shuffle downsampling (PD) with the blind spot network (BSN). 
Though PD can be utilized to meet the noise assumption of BSN, simply combining PD with CNN-based BSN is sub-optimal for dealing with spatially-correlated real noise. It causes damage to local details, thus bringing artifacts to the sub-sampled images, \textit{e.g.}, aliasing artifact, especially for large PD stride factors \cite{lee2022ap,zhou2020awgn}. 
Also, though more advanced designs of BSNs have been proposed \cite{wu2020unpaired,laine2019high,krull2020probabilistic}, CNN-based BSNs fail to capture long-range interactions due to their convolution operator, which is further bounded by the limited receptive field under the blind spot requirement.

In this paper, we present a novel method, called LG-BPN, to address these issues on self-supervised real image denoising, including the reliance on extra information, the loss of local structures
by noise correlation, 
and also the lacking of modeling distant pixel interaction. 
LG-BPN
can be directly trained without external information. 
Furthermore, we ease the destruction of fine textures by carefully considering the spatial correlation in real noise, at the same time injecting long-range interaction by tailoring Transformers to the blind spot network.
First, for local information, we introduce a densely-sampled patch-masked convolution (DSPMC) module. Based on the prior statistic of real noise spatial correlation, we take more neighbor pixels into account with a denser receptive field, allowing the network to recover more detailed structures.
Second, for global information, we introduce a dilated Transformer block (DTB). 
Under the special blind spot requirement, this greatly enlarges the receptive field compared with previous CNN-based BSNs, permitting more neighbors to be utilized when predicting the central blind spot pixel. 
These two designs enable us to fully exploit local and global information, respectively.
Extensive studies demonstrate that LG-BPN outperforms other state-of-the-art un-/self-supervised methods on real image denoising, as shown in Figure \ref{fig:sidd_val}.
We summarize our contributions as follows:
\begin{itemize}
	\item We present a novel self-supervised method called LG-BPN for real-world image denoising, which can effectively encode both the local detailed structure and the capture of global representation.

	\item Based on the analysis of real noise spatial correlation, we propose DSPMC module, which takes advantage of the higher sampling density on the neighbor pixels, enabling a denser receptive field for improved local texture recovery.

	\item To establish long-distance dependencies in previous CNN-based BSN methods, we introduce DTB, which aggregates global context while complying with the constraint of blind spot receptive field.
\end{itemize}

\section{Related Work}

\begin{figure*}[htbp]
  \centering
  \includegraphics[width=0.95\linewidth]{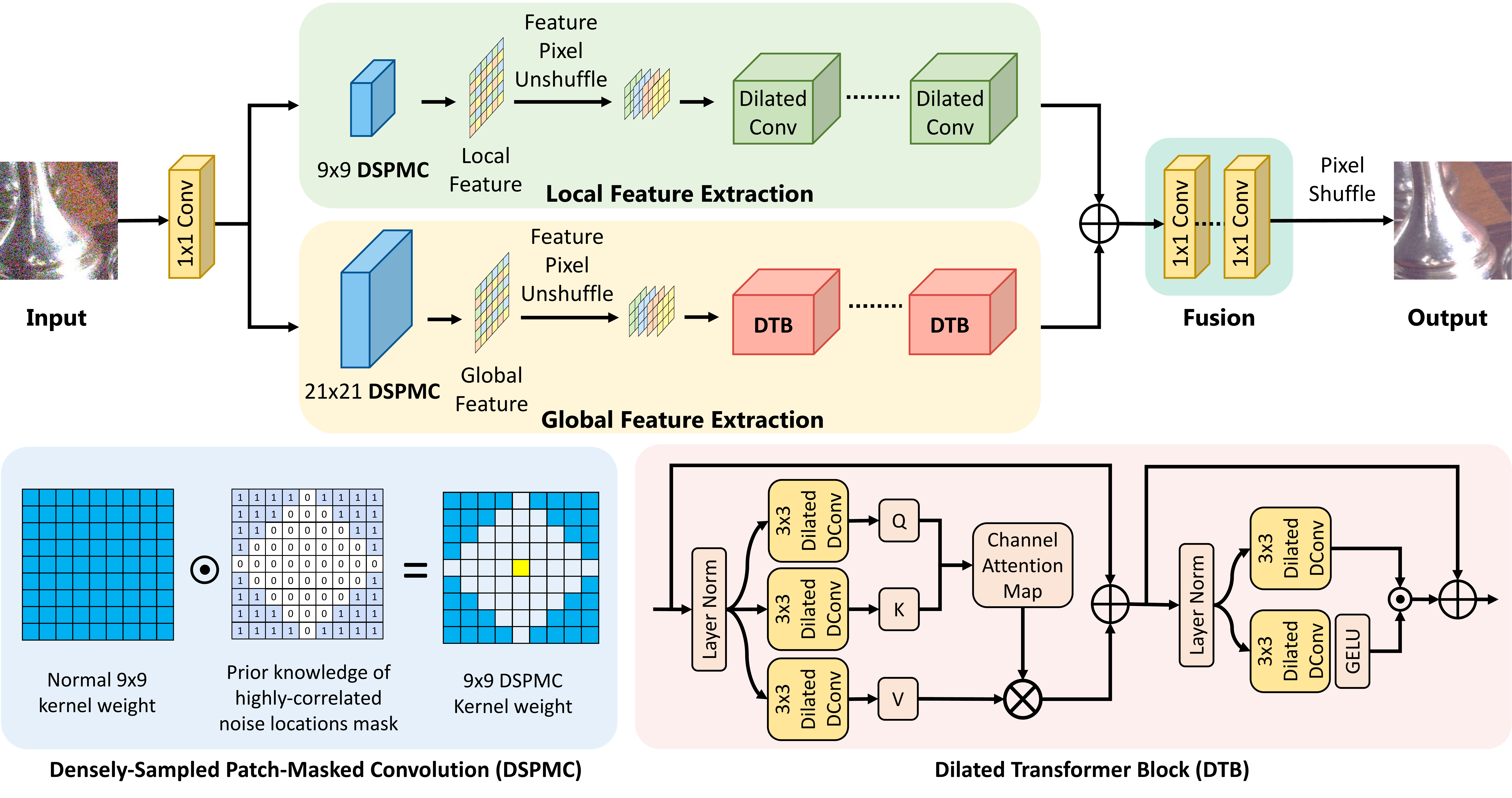}
  \vspace{-2mm}
  \caption{ The overall architecture of LG-BPN. Our method is composed of two branches, aiming at extracting local textures and global interactions, respectively. 
  For each branch, the input first goes through a DSPMC module, then further processed for the deep feature. Finally, the output of two branches is fused for the denoised result. \protect \footnotemark
  }
  \vspace{-5mm}
  \label{fig:overall_arch}
\end{figure*}

\subsection{Supervised Image Denoising}

DnCNN \cite{zhang2017beyond} is the first attempt to apply deep learning techniques to the image denoising task, where the training pairs are synthesized by additive white Gaussian noise (AWGN). Following DnCNN, several methods have been proposed for AWGN noise removal. For instance, FFDNet \cite{zhang2018ffdnet} advances it by taking the noise map as additional input. While achieving superior performance on AWGN removal, recent studies \cite{anwar2019real,guo2019toward} reveal the poor generalization ability of these models when applied to real noise, due to the gap between the noise distribution. 
The primary obstruction of real image denoising lies in the deficiency of real noisy-clean pairs. To this end, some real-world  denoising datasets are collected under carefully considered conditions \cite{abdelhamed2018high,brummer2019natural}. Based on these datasets, several methods \cite{guo2019toward,cheng2021nbnet} train the network directly on the real image pairs. Despite the decent performance, data collection can be extremely expensive and labor-intensive. Also, it is infeasible to collect clean images under complex scenes containing motion.

\subsection{Unsupervised Image Denoising}

Another line of research focuses on the situation where paired data is unavailable, including \textit{i)} generating pseudo noisy-clean image pairs, \textit{ii)} generating pseudo noisy-noisy image pairs, and \textit{iii)} training directly on noisy images.  

\noindent \textbf{Generating pseudo noisy-clean image pairs.} In situations where unpaired noisy-clean data is available, generation-based methods seek to synthesize real noise on clean images for aligned training data, which can be used for supervised methods.
Inspired by the generative adversarial network (GAN), GCBD \cite{chen2018image} synthesizes realistic noisy images to train the denoising network, while its performance is limited by the inaccurate consideration of noise components.
UIDNet \cite{wu2020unpaired} takes a step further by combining the distilled knowledge of the self-supervised denoising network and  extra information from synthetic pairs. C2N \cite{jang2021c2n} considers various noise components in real-world scenarios for more accurate noise synthesis. 
However, dealing with the gap between unpaired data is still challenging. The mismatch in scene distribution can result in inaccurate generation, thus degrading the quality of synthesized data.

\noindent \textbf{Generating pseudo noisy-noisy image pairs.}
The 
semi-supervised method Noise2Noise \cite{lehtinen2018noise2noise} uses multiple noisy images for training, 
which can be applied without clean images. 
However, the acquisition of multiple independent observations under the same scene is still less practical. 
Therefore, several methods seek to construct noisy-noisy pairs from a single noisy image. Neighbor2Neighbor \cite{huang2021neighbor2neighbor} generates two sub-sampled images under simplified noise assumptions. 
To handle complex noise distribution in real images, several methods have been proposed, including Noisier2Noise \cite{moran2020noisier2noise}, NAC \cite{xu2020noisy} and R2R \cite{pang2021recorrupted}. Still, these methods either require prior knowledge or are limited by specific constraints, which can be impractical in real situations. Specifically, Noisier2Noise \cite{moran2020noisier2noise} requires noise distribution when synthesizing noisy/noisy pairs. NAC \cite{xu2020noisy} works under the assumption that the noise level is relatively weak. R2R \cite{pang2021recorrupted} also uses additional information, \textit{e.g.}, noise level function (NLF) and image signal processing (ISP) function.

 \noindent \textbf{Training directly on noisy images.} 
Another type of method follows a self-supervised manner, which can be directly trained on the noisy images and free of synthesizing pseudo image pairs.
 Noise2Void \cite{krull2019noise2void} and Noise2Self \cite{batson2019noise2self} propose the self-supervised blind-spot strategy by masking the corresponding central pixel. 
Laine19 \cite{laine2019high} and D-BSN \cite{wu2020unpaired} are further proposed for advanced BSN designs, while the convolution-based architecture limits their exploitation for long-range dependencies. 
 To ease the information loss by the blind spot, Blind2Unblind \cite{wang2022blind2unblind} introduces a novel re-visible loss term. 
 Unfortunately, the above-mentioned methods work under the assumption that noise is pixel-wise independent, thus inevitably learning identity mapping under spatially-correlated real noise.
Towards real image denoising, CVF-SID \cite{neshatavar2022cvf} disentangles the noise components from the clean images,
but it assumes the noise is spatially-uncorrelated, which does not match the real noise distribution. Asymmetric pixel shuffle downsampling BSN (AP-BSN)\cite{lee2022ap} combines the pixel shuffle downsampling (PD) with a CNN-based BSN \cite{wu2020unpaired}. 
While achieving promising results, local structures are damaged by directly applying the PD operation to the image. 
Sub-sampled images are  corrupted by various artifacts, \textit{e.g.}, aliasing artifacts, which are more pronounced under a large PD stride factor \cite{lee2022ap,zhou2020awgn}. 
Also, adopting the CNN-based BSN leads to a limited receptive field. Since BSN recovers the central pixels based on its neighbors, fewer available neighbor pixels inevitably lead to performance loss.
In summary, this results in the inadequate utilization of information with respect to both local and global contexts.
Instead, our method benefits from a denser sampling density for improved local detail extraction, and also enjoys the distant pixel modeling ability for the enlarged receptive field.


\footnotetext{We use ‘global’ to differentiate from our ‘local’
branch. Though a
more accurate term is ‘non-local’, we follow the usage of
‘global’ as \cite{zamir2022restormer}.}
\vspace{-2mm}

\section{Method}

\vspace{-1mm}

We first illustrate the overall architecture of LG-BPN in Figure \ref{fig:overall_arch}, then elaborate on our motivation, and 
demonstrate our two core designs: DSPMC and DTB.

\begin{figure}[t]
  \centering
  \includegraphics[width=1\linewidth]{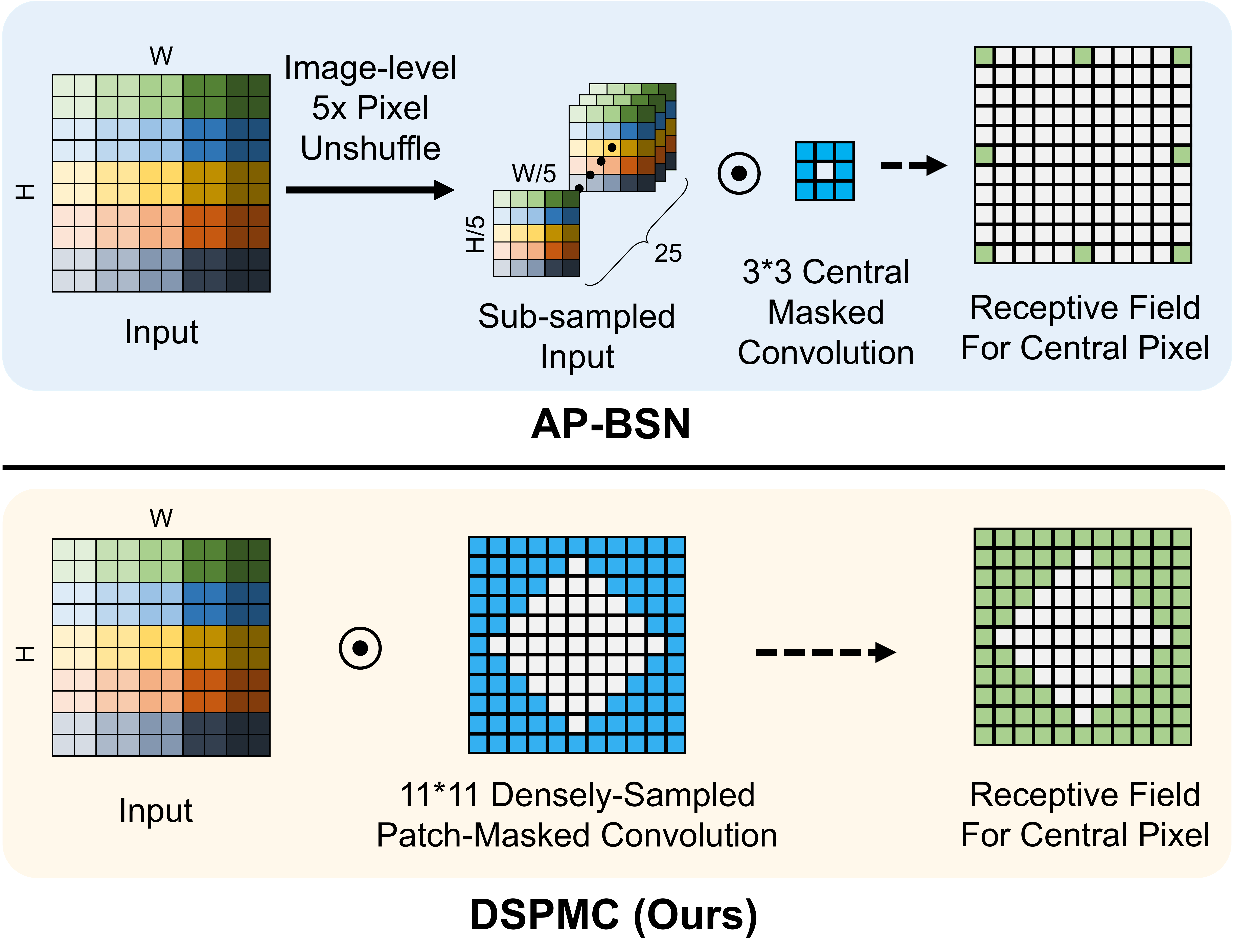}
  \vspace{-7.5mm}
  \caption{Comparison of receptive field for the central pixel of AP-BSN and our method. 
\textcolor[RGB]{102,204,50}{Green} pixels contribute to the restoration of central pixel.
\textcolor{blue}{Blue}
pixels represent convolution kernel. 
We realize a denser receptive field
by utilizing more neighbor pixels. \protect \footnotemark
}
  \label{fig:kernel}
  \vspace{-4mm}  
\end{figure}

\footnotetext{
We adopt 9$\times$9 DSPMC in the
local branch. Here, the 11$\times$11 size is 
shown 
for illustration purposes to better compare the receptive field.
}

\subsection{Motivation and Modeling}

Despite the decent results on simple synthetic noise removal, the performance of self-supervised denoising methods declines significantly when dealing with real noise, due to its strong spatial noise correlation.
This easily breaks the assumption on which most state-of-the-art methods are based, \textit{i.e.}, noise is pixel-wise independent. These methods
assume the clean signal of the central pixel is dependent on
neighbors, while the noise is independent instead. Thus under real scenarios, they inevitably misinterpret the spatially-dependent noise as clean signals, and fail to recover the underlying clean images.  
Consequently, careful consideration of the spatial correlation is a must for self-supervised real noise removal. While the existing methods either struggle with poor results when completely ignoring this correlation, or suffer details loss from ill consideration. 
For example, AP-BSN \cite{lee2022ap} meets the assumption of the powerful BSN by adopting PD on the input image. As shown in Figure \ref{fig:kernel}, though this breaks the spatial correlation, the 
sampling density is dramatically decreased by the PD. This severely degrades the extraction of fine details based on the Nyquist-Shannon sampling theorem, \textit{i.e.}, the fidelity of the results shows a positive correlation to sampling density.

Besides, though BSN is already adapted for state-of-the-art performance \cite{lee2022ap}, 
its potential is still heavily hampered by their inherent shortage, \textit{i.e.}, the limit on the receptive field by the masked pixels for avoiding identity mapping.
Despite
the recent advanced BSN designs
 \cite{wu2020unpaired,laine2019high}, CNN-based BSNs are still unable to fully address this issue due to their local convolution operator and fail to model the long-distance dependencies. This adversely affects the performance of BSN, as the number of neighbors around the blind spot used for inferring is dramatically reduced.

We aim to tackle both of these challenges in our methods. First, we realize better extraction for detailed textures from the perspective of sampling density. As shown in Figure \ref{fig:kernel}, DSPMC enables a denser sampling rate by leveraging more neighbor pixels, which raises the upper limit of reconstruction quality. Second, by tailoring normal Transformers to a blind fashion, DTB is introduced for its powerful global modeling ability to compensate for the limited receptive field of CNN-based BSNs. 

Based on these two modules, we now introduce the overall architecture of our method. As shown in Figure \ref{fig:overall_arch}, LG-BPN is mainly composed of two branches in parallel, aiming at local and global contexts reconstruction respectively. 
For the local feature extraction branch, we first apply the $9 \times 9$ DSPMC module. The densely extracted features are then down-sampled to break the spatial correlation.
Then, the feature maps go through dilated convolution with a dilation of 2.
For the global branch, the input image first goes through a $21 \times 21$ DMPMC module with a larger receptive field, which is then processed by DTBs. Finally, the local and global information from the two branches is fused together for the final output.

\begin{figure}[t]
\centering
\hfill
	\subcaptionbox{Bar plot of correlation by relative position \label{fig:corre_bar}}{\includegraphics[width = 0.22\textwidth]{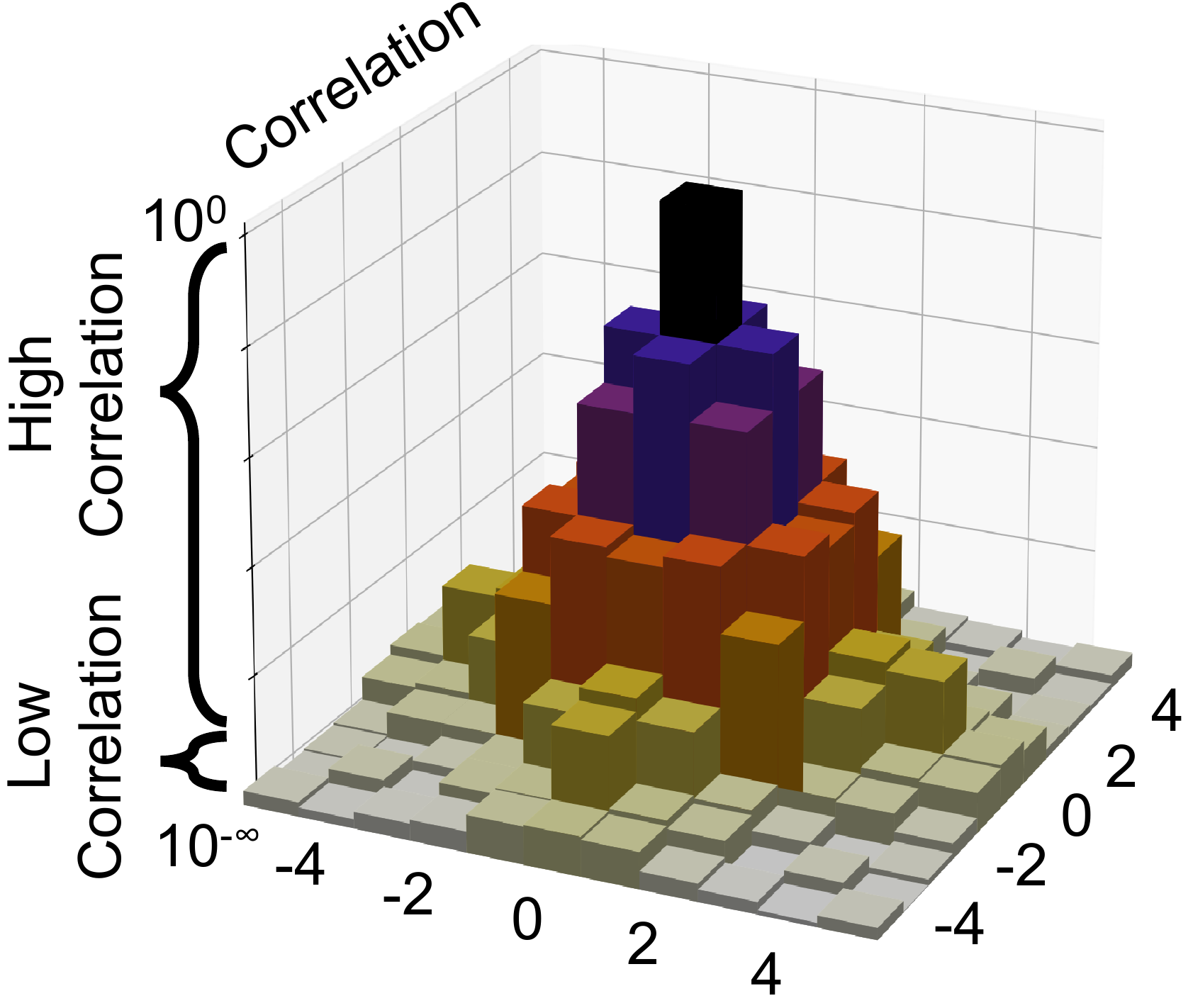}}
	\hfill
	\subcaptionbox{High-correlation pixels mask\label{fig:corre_mask}}{\includegraphics[width = 0.18\textwidth]{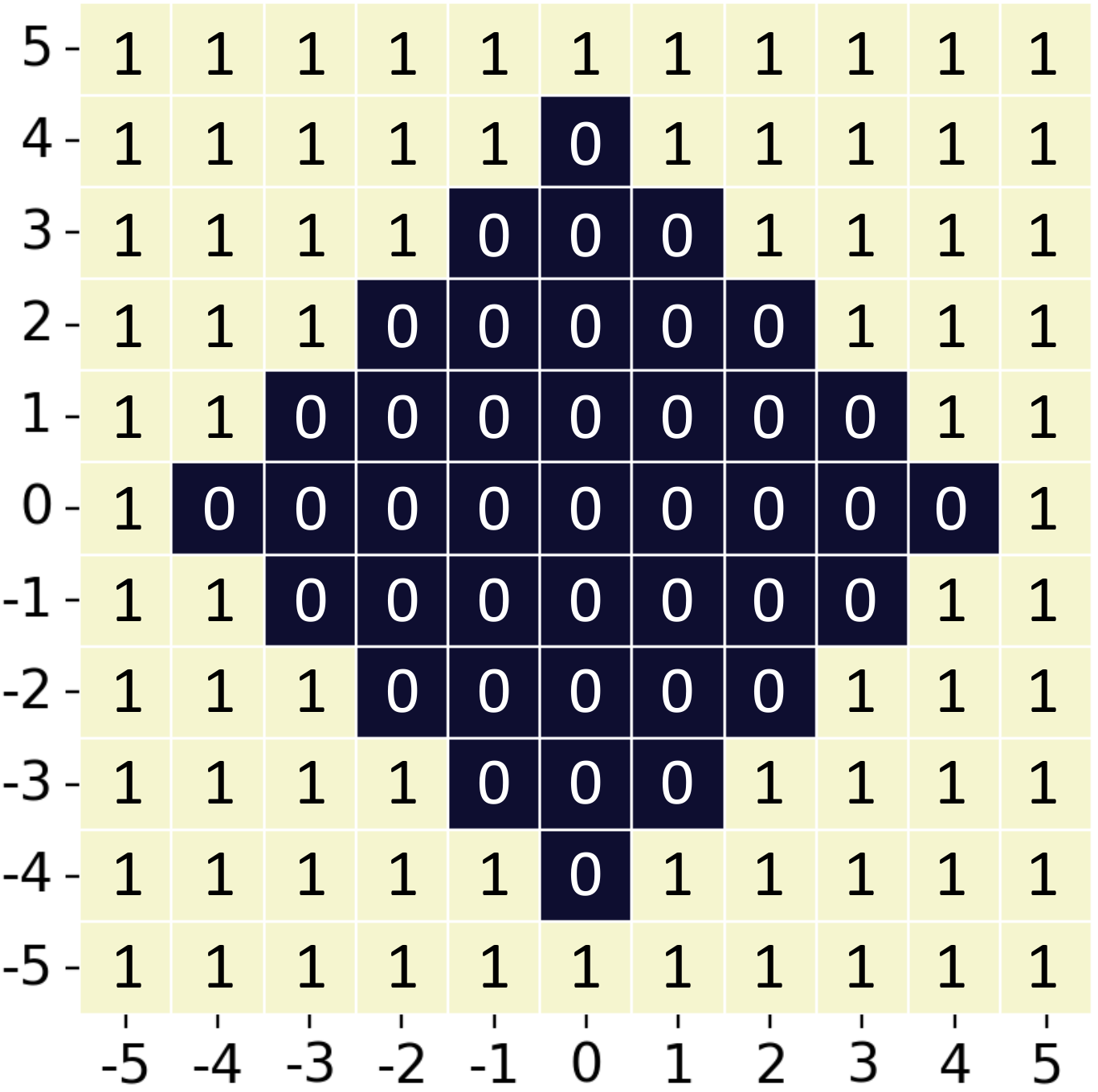}}
	\hfill
\vspace{-2mm}
\caption{Visualization of spatial correlation in real noise. (a) Bar plot of spatial correlation calculated by the correlation coefficient. Note that we scale the height by log norm for better visualization.
 The higher bar indicates a stronger correlation. (b) The mask of high-correlation pixels. Locations with 0 mean this pixel is strongly correlated to the central pixel, while 1 means not.}
\label{fig:correlation}
  \vspace{-6mm}
\end{figure}

\subsection{Densely-Sampled Patch-Masked Convolution}

Neither adopting a low sample rate nor sampling all neighbor pixels can be an optimal choice for BSN when tacking real noise.
In DSPMC, we aim to extract as much local information as possible, at the same time avoiding misinterpretation by these strongly-correlated neighbor pixels. To this end, we start by presenting the relationship between spatial correlation and the relative position, as shown in Figure \ref{fig:correlation}. Following previous works \cite{lee2022ap,zhou2020awgn}, we use Pearson's correlation coefficient to depict the relationship. Specifically, we first obtain the noise map by subtracting the clean images from the noisy images in SIDD medium \cite{abdelhamed2018high} dataset. Then, correlation coefficient can be calculated by:
\begin{equation}
\rho_{N_{cen}, N_{nei}}=\frac{\operatorname{cov}(N_{cen}, N_{nei})}{\sigma_{N_{cen}} \sigma_{N_{nei}}},
\label{eqa:relation}
\end{equation}
where $N_{cen}$ and $N_{cen}$ represent the noise of the central pixel and neighbor pixels respectively.
In Figure \ref{fig:corre_bar}, we find that there exist more neighbor pixels which can also be leveraged for prediction. These pixels are not strongly correlated to the center pixel, thus bringing useful information, instead of misinterpretation to BSN.
Then, the DSPMC kernel can be calculated as:
\begin{equation}
\vspace{-1mm}
\begin{aligned}
 \mathbf{K}_{DSPMC} =  \mathbf{K}_{n} \odot  \mathbf{Mask}_{cor},
\end{aligned}
\end{equation}
where $ \mathbf{K}_{DSPMC}$ is the kernel of DSPMC,  $ \mathbf{K}_{n}$ is the kernel of the normal convolution, and $ \mathbf{Mask}_{cor}$ is the mask for filtering out highly-correlated pixels shown in Figure \ref{fig:corre_mask}.
 By integrating this noise distribution prior to sampling locations, our module can effectively take more neighbor pixels while avoiding the strongly correlated pixels. 
 This enables the extracted feature to contain more fine details, and a denser receptive field as well. Also, as shown in Figure \ref{fig:kernel}, since the extracted high-dimension feature already gathers the rich local details, the subsequent feature-level PD can save more useful information compared with the previous image-level PD.

However, directly applying this module is not an optimal choice: \textit{i)} the inference stage requires more details compared with training, so directly using the same architecture can damage high-frequency details \cite{lee2022ap}, \textit{ii)} a large kernel can cause computational inefficiency. 

\begin{figure}[t]
\centering

	\subcaptionbox{Kernel location during training \label{fig:kernel_train}}{\includegraphics[width = 0.15\textwidth]{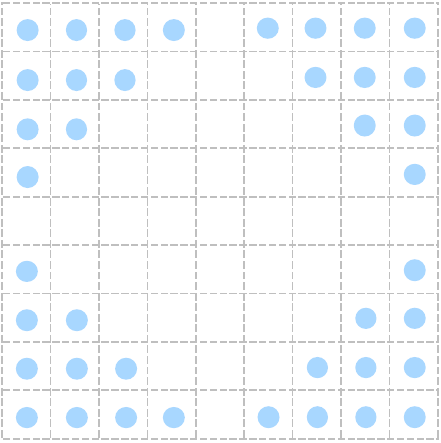}}
	\hfill
	\subcaptionbox{Shifted kernel location during testing\label{fig:kernel_test}}{\includegraphics[width = 0.15\textwidth]{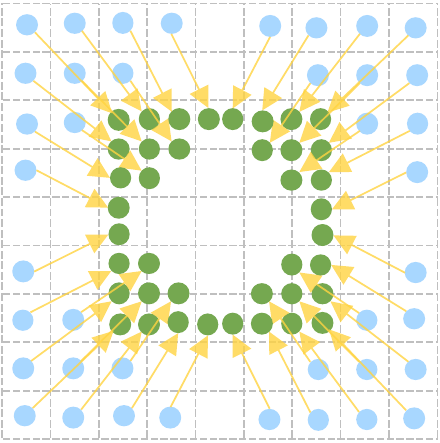}}
	\hfill
	\subcaptionbox{Shifted kernel location during testing with dilation\label{fig:kernel_test_dila}}{\includegraphics[width = 0.15\textwidth]{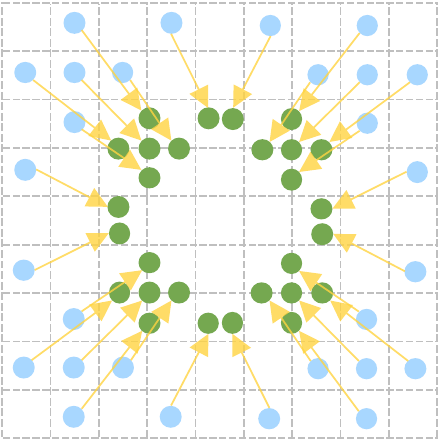}}

\vspace{-4mm}
\caption{Illustration of shifting kernel strategy and dilation of DSPMC. 
The \textcolor{blue}{blue} and \textcolor[RGB]{102,204,50}{green} points are the kernel locations during training and testing respectively.
\textcolor[RGB]{255,170,0}{Yellow} arrows indicate the shift direction. This 
avoids the strongly correlated pixels when training, sampling pixels closer to the center when testing for more details. 
}
\vspace{-4mm}
\label{fig:shift}
\end{figure}

\noindent \textbf{Kernel shift strategy.}
As shown in Figure \ref{fig:kernel_test}, for the first concern, 
we need to obtain more details while testing, focusing on local detailed information closer to the center pixel.
Inspired by the deformable convolution \cite{dai2017deformable}, we apply a set of fixed offsets on the kernel for each location, which enforce that the kernels are more gathered in the center:
\begin{equation}
\begin{aligned}
 \mathbf{y}(p_0) &= \sum_{k=1}^ \mathbf{K} w_k \cdot  \mathbf{x}\left(p_0+p_k+\Delta p_k\right) , \\
\Delta p_k  &= Ratio * (p_k - p_0),
\end{aligned}
\end{equation}
where $ \mathbf{K}$ is the kernel sampling locations, $w_k$ is the kernel weight, $ \mathbf{x}(p)$ and $ \mathbf{y}(p)$ denote the features at $p$ in input feature  $ \mathbf{x}$ and output feature $ \mathbf{y}$, and $\Delta p$ is the applied kernel offset. $Ratio$ is the extent we shift the kernel while testing.
By adding offsets to the kernel, we can \textit{shrink} the kernel and capture finer details while testing. 

\noindent \textbf{Dilation in DSPMC.}
 As shown in Figure \ref{fig:kernel_test_dila}, for the second concern, we further decrease the computational cost by adding dilation to the convolution kernel. This imposed sparsity makes our DSPMC computationally efficient especially for large kernel size, at the same time maintaining the dense receptive field for capturing detailed structures.

\begin{figure}[t]
\centering
	\subcaptionbox{\begin{tabular}[c]{@{}c@{}}Noisy input \end{tabular}}{\includegraphics[width = 0.145\textwidth]{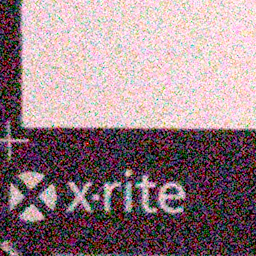}}
	\hfill
	\subcaptionbox{\begin{tabular}[c]{@{}c@{}}AP-BSN \cite{lee2022ap} \end{tabular}}{\includegraphics[width = 0.145\textwidth]{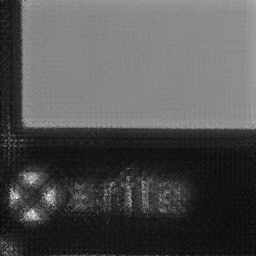}}
	\hfill
	\subcaptionbox{\begin{tabular}[c]{@{}c@{}} LG-BPN (Ours) \end{tabular}}{\includegraphics[width = 0.145\textwidth]{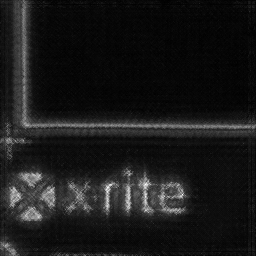}}
\vspace{-2mm}
\caption{The comparison of the feature map visualization of our method and AP-BSN \cite{lee2022ap} in the training phase. Our feature map shows clearer edges, validating the superiority of local details extraction of DSPMC by imposing denser sampling locations. }
  \label{fig:feature_vis}
  \vspace{-5mm}
\end{figure}

\noindent \textbf{Visualization of the extracted feature map.} 
To validate that our DSPMC module achieves a denser receptive field and is thus better at extracting local high-frequency structures, we present the visualization of the feature map. We select the output feature of the local extraction branch, and the corresponding location in AP-BSN. All channels are averaged and normalized for visualization. As shown in Figure \ref{fig:feature_vis}, in AP-BSN \cite{lee2022ap}, the local fine texture is damaged due to the insufficient use of neighbor signals. Instead, by leveraging more neighbor pixels, our feature map shows shaper edges and preserves more details.

\subsection{Dilated Transformer Block}

The receptive field of BSN is restricted by the imposed blind spots, while the local operator in CNN-based BSNs further prevents it from gathering global interaction. However, under the special blind spot constraint on the receptive field, it is non-trivial to directly introduce normal Transformer blocks to the BSN. Inspired by the D-BSN \cite{wu2020unpaired}, we aim to design the Transformer block without information exchange between spatially-adjacent pixels, which satisfies the blind spot requirement when combined with DSPMC. Under these requirements, we carefully consider the design of two core components in the Transformer: the self-attention calculation and the feed-forward layer. 

First, for the self-attention layer, spatial-wise attention enables spatial information exchange and thus does not meet our receptive field requirement. Recently, a grid-like self-attention can meet our requirement \cite{tu2022maxvit}, while the grid pattern
further narrows the receptive field that the blind spot has reduced.
 Instead, we adopt channel-wise attention \cite{zamir2022restormer} for its unawareness of spatial location
 and global perception as well.
  Furthermore, to enhance the local context while preventing information of adjacent pixels, we introduce dilated depth-wise convolution before computing feature similarity. For the input feature $\mathbf{X}$, the Query ($\mathbf{Q}$), Key ($\mathbf{K}$) and Value ($\mathbf{V}$) matrix is thus calculated by
$\mathbf{Q} = g^Q{(\mathbf{X})}$, $\mathbf{K} = g^K{(\mathbf{X})}$, $\mathbf{V} = g^V{(\mathbf{X})}$, where $g^Q(\cdot)$, $g^K(\cdot)$ and $g^V(\cdot)$ denote the dilated $3\times3$ depth-wise convolution.
Given the $\mathbf{Q}$, $\mathbf{K}$ and $\mathbf{V}$ matrix, the channel interaction can be obtained by the dot-product, where the attention map is of size $\mathbb{R}^{{C} \times {C}}$, and ${C}$ is the number of channels. The overall self-attention layer is represented as:
\begin{equation}
\vspace{-1mm}
\begin{aligned}
 \operatorname{Attention}(\mathbf{Q}, \mathbf{K}, \mathbf{V}) = & \mathbf{V}  \operatorname{Softmax}(\mathbf{K}  \mathbf{Q}), \\
 \mathbf{\hat{X}} = \operatorname{Attention}&(\mathbf{Q}, \mathbf{K}, \mathbf{V}) + \mathbf{X},
\end{aligned}
\vspace{-1mm}
\end{equation}
where $\mathbf{X}$ and $\mathbf{\hat{X}}$ denote the input and output features.

\begin{figure}[t]
\centering
\hfill
	\subcaptionbox{AP-BSN \cite{lee2022ap}}{\includegraphics[width = 0.16\textwidth]{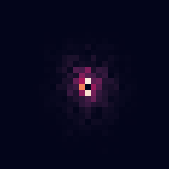}}
	\hfill
	\subcaptionbox{LG-BPN (Ours)}{\includegraphics[width = 0.16\textwidth]{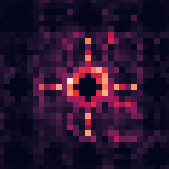}}
	\hfill 
\vspace{-2mm}
\caption{The comparison of the receptive field between AP-BSN \cite{lee2022ap} and our method. We calculate the gradient response to the central pixel. The brighter color represents the higher contribution for recovering the central pixel.}
  \label{fig:receptive_field}
  \vspace{-4mm}
\end{figure}

Second, for the feed-forward layer, adjacent information exchange can be simply avoided by adopting $1\times1$ convolutions only. Nonetheless, this fails to capture local context, which can be critical for restoring high-frequency details. We address this issue by also introducing dilation into the normal $3\times3$ convolution in the feed-forward layer. Then, features extracted by dilated depth-wise convolution go through a gating unit for the non-linearity. This gating unit is the element-wise product of two parallel paths, with one of them activated by the GELU unit.  The overall process of the feed-forward layer is formulated as:
\begin{equation}
\vspace{-1mm}
\begin{aligned}
\mathbf{G^1} &= g^1 ( \mathrm{LN}(\mathbf{X}) ) , \\
\mathbf{G^2} &= g^2(\mathrm{LN}(\mathbf{X}) ), \\
\mathbf{\hat{X}} &=  \operatorname{GELU}{(\mathbf{G^1}) \odot \mathbf{G^2}} + \mathbf{X},
\\
\end{aligned}
\vspace{-1mm}
\end{equation}
where $ \odot$ is element-wise multiplication, LN denotes layer normalization, $g^1(\cdot)$ and $g^2(\cdot)$ represent the $3\times3$ dilated depth-wise convolution. An additional benefit is that, compared to the normal $3\times3$ convolution, the introduced dilation can also enlarge the receptive field.

To prove the effectiveness of the introduced global dependencies, we also plot the receptive field for recovering the central pixel of our method and the CNN-based BSNs in Figure \ref{fig:receptive_field}. By injecting the long-range interaction into the blind spot network, more neighbor pixels are activated for restoring the central pixels in our method, offering a broader receptive field compared to the previous CNN-based BSNs.

\begin{table*}
\small
  \centering
\setlength{\tabcolsep}{3.1mm}{
 \renewcommand{\arraystretch}{0.9}{
\begin{tabular}{cclcccc}

                \toprule
                    
                \rowcolor[rgb]{ .949,  .949,  .949}
                &    &  & \multicolumn{2}{c}{SIDD} & \multicolumn{2}{c}{DND} \\
                \rowcolor[rgb]{ .949,  .949,  .949}
                 \multirow{-2}{*}{Type of supervision} & \multirow{-2}{*}{Training data} & \multirow{-2}{*}{Method}                                   & PSNR & SSIM & PSNR & SSIM                      \\
               
               \midrule

\multirow{2}{*}{Non-learning based}  & \multirow{2}{*}{None}                 & BM3D \cite{dabov2007image}                   & 25.65                     & 0.685                     & 34.51                     & 0.851                     \\
                                     &                                       & WNNM \cite{gu2014weighted}                   & 25.78                     & 0.809                     & 34.67                     & 0.865                     \\

\midrule

\multirow{8}{*}{Supervised}          & \multirow{3}{*}{Synthesized Pairs}    & DnCNN \cite{zhang2017beyond}  & 23.66                     & 0.583                & 32.43                     & 0.790                   \\
                                     &                                       & CBDNet \cite{guo2019toward} & 33.28                     & 0.868                     & 38.05                     & 0.942                     \\
                                     &                                       & 
AWGN-M \cite{zhou2020awgn} 
                                     &  \hspace{0.004\linewidth} 33.99$^\star$              & \hspace{0.008\linewidth}0.896$^\star$            & 38.40         & 0.945           \\
                                     \cmidrule[0.6pt]{2-7}
                                     & \multirow{5}{*}{Real pairs}           & DnCNN \cite{zhang2017beyond} & \hspace{0.008\linewidth}35.34$^\star$                     & \hspace{0.008\linewidth}0.885$^\star$                     & \hspace{0.008\linewidth}37.83$^\star$                     & \hspace{0.008\linewidth}0.929$^\star$                     \\
                                     &                                       & AINDNet \cite{kim2020transfer} & 38.84                     & 0.951                     & 39.34                     & 0.952                     \\
                                     &                                       & RIDNet \cite{anwar2019real}            & 38.70                     & 0.950                     & 39.25                     & 0.952                     \\
                                     &                                       & DIDN \cite{yu2019deep}           & 39.82                     & 0.973                     & 39.62                     & 0.954                     \\
                                     
                                     \midrule
                                     
\multirow{11}{*}{Unsupervised}        & \multirow{4}{*}{Noisy-clean pairs} & GCBD \cite{chen2018image}         & -                         & -                         & 35.58                     & 0.922                     \\ 
                                     &                                       & UIDNet \cite{hong2020end}           & 32.48                     & 0.897                     & -                         & -                         \\
                                     &                                       & C2N \cite{jang2021c2n} + DIDN \cite{yu2019deep}             & 35.35                     & 0.937                     & 36.38                     & 0.887                     \\
                                     &                                       & D-BSN \cite{wu2020unpaired} + MWCNN \cite{liu2018multi}          & -                         & -                         & 37.93                     & 0.937                     \\

                                     \cmidrule[0.6pt]{2-7}
                                                               
     & \multirow{3}{*}{Noisy-noisy pairs}   & Noise2Self \cite{batson2019noise2self} & \hspace{0.008\linewidth}29.56$^{\dagger}$                     & \hspace{0.008\linewidth}0.808$^{\dagger}$                     & -                         & -                         \\
                                     &                                       & NAC \cite{xu2020noisy}            & -                         & -                         & 36.20                     & 0.925                     \\
                                     &                                       & R2R \cite{pang2021recorrupted}          & 34.78                     & 0.898                     & -                         & -                         \\
                                     
                                      \cmidrule[0.6pt]{2-7}
                                     
                                     & \multirow{4}{*}{\begin{tabular}[c]{@{}c@{}}Single noisy \\observation\end{tabular}}         & Noise2Void \cite{krull2019noise2void}         & \hspace{0.008\linewidth}27.68$^{\dagger}$                     & \hspace{0.008\linewidth}0.668$^{\dagger}$                     & -                         & -                         \\
                                     &                                       & CVF-SID \cite{neshatavar2022cvf} & 34.71                     & 0.917                     & 36.50                     & 0.924                     \\
                                     &                                       & AP-BSN \cite{lee2022ap}           & 35.97                     & 0.925                     & 38.09                     & 0.937                     \\
                                     &                                       & \textbf{LG-BPN (Ours)}                  & \textbf{37.28}                     & \textbf{0.936}                     & \textbf{38.43}                     & \textbf{0.942}                    
\\
\bottomrule
\\
\end{tabular}
}
}
\vspace{-4mm}
  \caption{Quantitative comparison of various methods on SIDD and DND benchmark datasets. 
  Though several supervised methods achieve better results using noisy/clean image pairs, our methods use noisy RGB images only. 
  Results with $\star$ mean these are reproduced and evaluated by ourselves, since they are not evaluated on the dataset we use in their original paper. The results marked with ${\dagger}$ are reported from R2R \cite{pang2021recorrupted}. Otherwise, we report the official results from SIDD and DND benchmark websites.}
  \vspace{-4mm}
  \label{tab:main}
\end{table*}

\section{Experiments}

\subsection{Dataset and Setup Details}

We train and evaluate our method on two real-world datasets, \textit{i.e.}, SIDD \cite{abdelhamed2018high} and DND \cite{plotz2017benchmarking}. Note that for the SIDD benchmark dataset and DND benchmark dataset, we submit the output to the website for online evaluation. 

\noindent \textbf{Smartphone Image Denoising Dataset (SIDD) \cite{abdelhamed2018high}} contains paired images for real-world denoising by five smartphone cameras. For training, we use the sRGB images from SIDD-Medium including 320 pairs. For validation and evaluation, we use the sRGB images from the SIDD validation set and benchmark set respectively. Each includes 1280 patches of size $256\times256$, where the ground truth images are also provided for the validation set.

\noindent \textbf{Darmstadt Noise Dataset (DND) \cite{plotz2017benchmarking}} contains 50 noisy images for benchmarking without the ground truth provided, and the results can only be obtained via the online submission system. Therefore, we enjoy a fully self-supervised manner and directly train our method on the test set without extra eternal data.

\subsection{Training Details}

During training, we keep the same setting as the previous work \cite{lee2022ap}. Specifically, a batch size of 8 is used in the experiment. We adopt $\ \mathcal{L}_1$ loss between ground truth and output for training. The learning rate starts with 
$1e$-4, where Adam
optimizer is adopted. The network is trained with 20 epochs until it fully converges. We implement the method in PyTorch 1.8.0, and train our model on the Nvidia RTX 3090.
Two metrics are utilized to evaluate the performance of methods, including peak signal-to-noise ratio (PSNR) and structural similarity (SSIM) \cite{wang2004image}. The larger value of PSNR and SSIM implies better fidelity.

\subsection{Evaluation of Real-world Denoising}

\begin{figure*}
 \centering
\small
 \setlength{\tabcolsep}{0.115cm}
\hspace{1mm}
\begin{minipage}[l]{0.288\linewidth}
\begin{flushleft}
 \vspace{-0.1mm}
  \hspace{-20mm}
\includegraphics[width=0.95\linewidth]{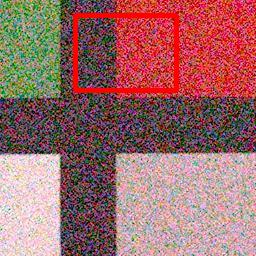}
   \begin{tabular}[c]{@{}c@{}}\hspace{-8.4mm}SIDD benchmark input\end{tabular}
  \end{flushleft}
 \end{minipage}
 \hspace{-22mm}
 \begin{minipage}[t]{0.6\linewidth}
  \begin{tabular}{cccc}
   \includegraphics[width=0.275\linewidth]{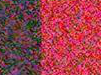}
  & \includegraphics[width=0.275\linewidth]{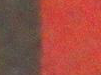}
  & \includegraphics[width=0.275\linewidth]{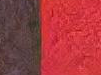}
  &\includegraphics[width=0.275\linewidth]{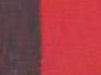}
   \\
   
    N2V \cite{krull2019noise2void} & DnCNN \cite{zhang2017beyond} &  AWGN-M \cite{zhou2020awgn} & C2N \cite{jang2021c2n} \\ 

 \includegraphics[width=0.275\linewidth]{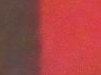}
  &\includegraphics[width=0.275\linewidth]{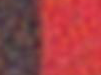}
  &\includegraphics[width=0.275\linewidth]{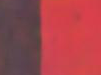}
  &\includegraphics[width=0.275\linewidth]{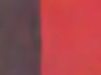}
  
  \\
  R2R \cite{pang2021recorrupted} &  CVF-SID \cite{neshatavar2022cvf} &  AP-BSN \cite{lee2022ap} & LG-BPN (Ours) \\
  \end{tabular}
 \end{minipage}

 \vspace{-1mm}
 \caption{Visual quality comparison on SIDD benchmark dataset. Note that the quantitative results are not available.}
    \vspace{-1mm}
 \label{fig:SIDD_bench} 
\end{figure*}

We validate the effectiveness of our method for real-world image denoising on the commonly-used SIDD benchmark dataset and DND benchmark dataset. 
Table \ref{tab:main} shows the comparison of various methods on SIDD and DND benchmark datasets.  
Visualization results of several methods addressed in Table \ref{tab:main} on SIDD and DND datasets can be found in Figure \ref{fig:SIDD_bench} and Figure \ref{fig:DND}. 
We achieve better results in quantitive and qualitative metrics than previous un-/self-supervised methods.
Compared with unsupervised methods trained on unpaired clean-noisy data, LG-BPN does not rely on extra data for synthesizing training pairs, also avoiding the misalignment between the scene distribution. For the self-supervised methods, NAC \cite{xu2020noisy} works under the weak noise level assumption, while R2R \cite{pang2021recorrupted} can not be directly applied to sRGB images without extra NLF and ISP functions, both of which harm their performance in real-world situations. In contrast,  LG-BPN can be directly applied and not restricted by these assumptions. For methods leveraging single noisy observations, CVF-SID \cite{neshatavar2022cvf} does not consider the strong spatial correlation property in real noise, thus real noise can not be fully removed as shown in Figure \ref{fig:SIDD_bench}. AP-BSN \cite{lee2022ap} suffers from inadequate sample locations with a limited receptive field, so details are blurred as shown in Figure \ref{fig:DND}. 
Instead, LG-BPN carefully integrates the spatial correlation into the network design, simultaneously modeling distant context dependencies.

\begin{table*}
    \centering
    \begin{subtable}[t]{0.57\linewidth}
         \centering
 \resizebox{\columnwidth}{!}{
 \setlength{\tabcolsep}{2mm}
 \renewcommand{\arraystretch}{1}
  \begin{tabular}{cccccc}
   \bottomrule[0.8pt]
  \rowcolor[rgb]{ .949,  .949,  .949}   \begin{tabular}[c]{@{}c@{}}9x9 DSPMC \\ dilation \end{tabular}    & \begin{tabular}[c]{@{}c@{}}21x21 DSPMC \\ dilation\end{tabular}    & PSNR    & SSIM    &  FLOPS (G) & Params (M)   \bigstrut\\
   \hline
 1 & 1  &   37.23  & 0.885   & 88.6  & 1.35   \bigstrut[t]\\
 \textbf{1} &  \textbf{2}  &    \textbf{37.32} &  \textbf{0.886}    &  \textbf{29.8}  &  \textbf{0.45}  \\
 1 &3  &   37.23 & 0.883    & 17.1  & 0.26 \\
 2 &1  &   36.85 & 0.879    & 84.6  & 1.23 \\
 2 &2  &   36.84 & 0.875    & 25.8  & 0.39 \\
 2 &3  &   36.99 & 0.873    & 13.1  & 0.20 \bigstrut[b]\\

        \toprule[0.8pt]
    \end{tabular}%
     
    }
\caption{Ablation studies on dilation rates. Different dilation rate combinations are explored on the DSPMC in the local branch and global branch.}
\label{tab:dilation}
       
    \end{subtable}
    \hspace{3.1mm}
    \begin{subtable}[t]{0.4\linewidth}
      \centering
 \scriptsize
 \resizebox{\columnwidth}{!}{
  \setlength{\tabcolsep}{2mm}
  \renewcommand{\arraystretch}{1}
    \begin{tabular}{ccccc}
    \bottomrule[0.6pt]
    \rowcolor[rgb]{ .949,  .949,  .949}  Method  & PSNR & SSIM \bigstrut\\
    \hline
     w/o 9$\times$9 DSPMC &   35.82  &    0.855  \bigstrut[t]  \\
 w/o 21$\times$21 DSPMC &   36.21  &     0.869       \\
Replacing Conv with DTB &  36.90   &    0.880        \\
Replacing DTB with Conv&  36.82   &    0.875        \\
 w/o kernel shift &   35.64  & 0.857 \\ 
 \textbf{LG-BPN (Ours)} &   \textbf{37.32}   &    \textbf{0.886} \bigstrut[b] \\
        \toprule[0.6pt]   
    \end{tabular}%

    }    \caption{Ablation studies on our proposed method. Improvements can be found with our proposed modules and network design.}
    \label{tab:structure}
    \end{subtable}
    \vspace{-2mm}
 \caption{The analysis of our method on the SIDD validation dataset. Experimental results prove the effectiveness of our method design.}
     \vspace{-3mm}
 \label{tab:ablation}
\end{table*}

\begin{figure}
    \centering
     \scriptsize
    \setlength{\tabcolsep}{0.05cm}
   \hspace{-3mm}
   \begin{minipage}[l]{0.29\linewidth}
    \vspace{-3mm}
   \begin{center}
   \includegraphics[width=1.24\linewidth]{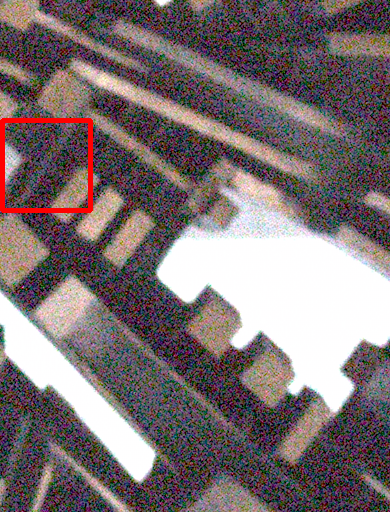}
   \begin{tabular}[c]{@{}c@{}}\hspace{3mm}DND benchmark input\end{tabular}
   \end{center}
    \end{minipage}
    \hspace{5mm}
    \begin{minipage}[t]{0.6\linewidth}
     \begin{tabular}{ccc}
    \includegraphics[width=0.33\linewidth]{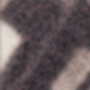}
     & \includegraphics[width=0.33\linewidth]{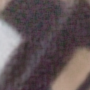}
     & \includegraphics[width=0.33\linewidth]{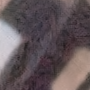}
     \\
     NAC   \cite{zhang2017beyond}     & C2N \cite{jang2021c2n} &  AWGN-M \cite{zhou2020awgn} \\
      28.50/0.777 & 29.20/0.793 & 29.26/0.797\\
      
     \includegraphics[width=0.33\linewidth]{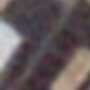}
       &\includegraphics[width=0.33\linewidth]{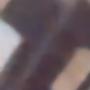}
         &\includegraphics[width=0.33\linewidth]{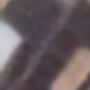}
      
        \\ CVF-SID \cite{neshatavar2022cvf} & AP-BSN \cite{lee2022ap} & \textbf{LG-BPN (Ours)} \\ 
        30.22/0.855 & 32.94/0.889 & \textbf{{33.37}}/\textbf{{0.897}}

   \\
     \end{tabular}
    \end{minipage}
   
    \vspace{-2mm}
    \caption{Visual quality comparison on DND benchmark dataset.}
    \vspace{-5mm}
    \label{fig:DND} 
   \end{figure}


\subsection{Analysis of the Proposed Method}

\noindent \textbf{Dilation factor in densely-sampled convolution.} Directly introducing densely-sample convolution can be computationally expensive. To balance the efficiency and the performance, we further introduce the sparsity to the convolution kernel by adding dilation to the original DSPMC. Figure \ref{fig:kernel_test_dila} shows the illustration of the dilation.

To explore the better trade-off between performance and efficiency, we provide ablation studies on the dilation rate for our DSPMC in both local and global extraction branches. As shown in Table \ref{tab:dilation}, a dilation of 1 for $9\times9$ DSPMC and 2 for $21\times21$ DSPMC achieve a better balance. We claim its reason is that the difference in kernel size results in focusing on varied scales of information. This imposes different sensitivity to the dilation, \textit{i.e.}, sampling density. For relatively small $9\times9$ kernels, it focuses more on the local textures, thus adding dilation can notably lower its ability when reconstructing detailed structures. While for the $21\times21$ kernel, the larger kernel size makes it aim at the global context more. Thus, the introduced dilation does not severely harm its global extraction ability, at the same time reducing the computational cost.

\noindent \textbf{The exploitation of local and global information.}
LG-BPN consists of two branches in parallel.
Specifically, we combine the small $9\times9$ DSPMC with dilated convolution and the large $21\times21$ DSPMC with DTB, focusing on local and global context processing respectively. 
To validate the reasonableness of our network architecture, we conduct ablation studies on these components. Table \ref{tab:structure} shows the results of different combinations.

It can be seen that removing the DSPMC of size $9\times9$ and $21\times21$ in either branch severely degrades the performance. It is on account of the insufficient sampling density, which limits the reconstruction quality according to the Nyquist-Shannon sampling theorem. In this situation, the sampling density is made severely sparse, which causes insufficient utilization of input signals. This performance drop demonstrates the effectiveness of our DSPMC module.
 
Furthermore, we validate that the different sizes of DSPMC in local and global feature extraction branches make them focus on various scales of features. Such discrepancy across the scale demands us to treat scale-speciﬁc characteristics in a different way. When replacing the dilated convolution in the local branch with our DTB, the lack of local connectivity brings inferior performance.
Similarly, when replacing the DTB in the global branch with the dilated convolution, the network is built by convolutions only. It induces a lack of long-range interaction, which greatly limits their recovery quality as well. This proves the superior design of our architecture. By exploiting the locality of convolution with a small DSPMC, also the global dependencies of DTB with a large DSPMC, our architecture enjoys more reasonable exploitation for multi-scale context.

\noindent \textbf{The effect of kernel shift strategy.} Since the requirement for pixel-wise independent noise is different in training and testing, directly applying the same kernel of the training phase while testing will lose image details \cite{lee2022ap}. We proposed a kernel shift strategy, as illustrated in Figure \ref{fig:kernel_test}. In Table \ref{tab:structure}, the lack of kernel shift causes 1.68 dB drops, proving the effectiveness of our shifting paradigm.

\vspace{-2mm}
\section{Conclusion}
\vspace{-1mm}

In this paper, we propose
LG-BPN for self-supervised real image denoising, aiming to address the 
details lost by the coarse consideration for real noise
correlation, and the lack of global interaction by the inherent constraint on the receptive field for BSN. First, we propose DSPMC to fully preserve the local structures. 
Owing to
a denser receptive field, we ease the destruction of fine textures and can thus better reconstruct
details.
Second, we propose DTB, injecting
distant
interactions into the previously CNN-based blind spot networks.
Since blind spot networks rely on neighbor signals for predicting, more clues can be provided by activating more neighbor pixels. 
Extensive results on real-world 
datasets reveal
the superior performance of LG-BPN.

\vspace{1mm}
\noindent 
\textbf{Acknowledgement} This work was supported by the National Natural Science Foundation of China under Grants No. 62171038, No. 61827901, No. 62088101, and No. 62006023.


{\small
\balance
\bibliographystyle{ieee_fullname}
\bibliography{egbib}
}

\end{document}